# Yuan 2.0-M32: Mixture of Experts with Attention Router

Shaohua Wu*, Jiangang Luo, Xi Chen, Lingjun Li, Xudong Zhao, Tong Yu, Chao Wang,
Yue Wang, Fei Wang, Weixu Qiao, Houbo He, Zeru Zhang, Zeyu Sun, Junxiong Mao, Chong Shen

IEIT Systems

**ABSTRACT**

Yuan 2.0-M32, with a similar base architecture as Yuan-2.0 2B, uses a mixture-of-experts architecture with 32 experts of which 2 experts are active. A new router network, Attention Router, is proposed and adopted for a more efficient selection of experts, which improves the accuracy compared to the model with classical router network. Yuan 2.0-M32 is trained with 2000B tokens from scratch, and the training computation consumption is only 9.25% of a dense model at the same parameter scale. Yuan 2.0-M32 demonstrates competitive capability on coding, math, and various domains of expertise, with only 3.7B active parameters of 40B in total, and 7.4 GFlops forward computation per token, both of which are only 1/19 of Llama3-70B. Yuan 2.0-M32 surpass Llama3-70B on MATH and ARC-Challenge benchmark, with accuracy of 55.89 and 95.8 respectively. The models and source codes of Yuan 2.0-M32 are released at Github[1].

## 1. Introduction

Given a fixed amount of computation for each token, a model with Mixture of Experts (MoE) structure can be easily built on a much larger scale than a dense model by increasing the number of experts, and thus achieves a higher accuracy performance. In reality, it is common to train a model with limited computational resource, and the MoE is considered as a good candidate to reduce the substantial cost associated with the extreme large scale of model, datasets and limited computing power.

The idea of MoE dates back to 1991 (Jacobs et al., 1991). The total loss is the combination of weighted loss of each expert with the ability to make independent judgement. The concept of sparsely-gated MoE has been first brought into focus by Shazeer et al. (2017) in a translation model. With this routing strategy, a very small number of experts will be active for reasoning instead of calling all experts simultaneously. This sparsity also allows the model to scale to 1000 times between stacked LSTM layers at the expense of very little computational efficiency. The Noisy Top-K Gating routing network introduces some adjustable noise to the softmax function and keeping the top-K value, to balance expert utilization. In recent years, with the ever-increasing model size, the role of routing strategy has attracted more attention for efficient allocation of computation resources.

The experts routing network is the core in a MoE structure. This structure selects candidate experts to participate in the computation by calculating the probability of token allocation to each expert. Currently, in most popular MoE structures, it is common to adopt a classical routing algorithm that performs a dot product between the token and the feature vector representing each expert, and then selects the experts with the largest dot product value (Shazeer et al. 2017; Fedus, Zoph and Shazeer, 2022; Zhou et al., 2022). The feature vectors of the experts in this transformation are independent, ignoring the correlation between experts. However, the MoE structure usually select more than one expert each time, and multiple experts often participate in calculation collaboratively, which means there should be an inherent correlation between experts. It will undoubtedly improve the accuracy of the model, if the relationship between experts is considered in the process of selecting experts.

*wushaohua@ieisystem.com
[1]https://github.com/IEIT-Yuan/Yuan2.0-M32

The major contribution of our work are summarized as follows:
1) Propose the Attention Router that considers the correlation between experts, resulting in a higher accuracy compared with the classical router structure.
2) Release the Yuan 2.0-M32 model with 40B total parameters and 3.7B active ones. There are 32 experts in total and 2 experts activated for each token. The computational consumption for training is only 1/16 of that for a dense model at a similar parameter scale, and the cost for inference is similar to a dense model with 3.7B parameters.

## 2. Related Works

Gshard (Lepikhin et al., 2020), a giant model with over 600 billion parameters, introduces the MoE method into Transformer Encoder for the first time, and provides an efficient distributed parallel computing architecture with routing across accelerators. Switch Transformer (Fedus, Zoph and Shazeer, 2022) simplifies the MoE routing algorithm with sparse routing. Zhou et al. (2022) has proposed a new MoE routing algorithm called Expert Choice (EC) routing algorithm to achieve the optimal load balancing in the MoE system. Mistral 8x7B model surpasses model with 10 times larger parameters in several human benchmarks with classical routing network (Jiang et al., 2024). DBRX uses a fine-grained MoE architecture and chooses 4 experts among 16 (Mosaic AI research, 2024). DeepSeekMoE improves the expert specialization with fine-grained expert segmentation as well as shared expert isolation (Dai et al., 2024). The shared experts activate tokens for all inputs and are not affected by the routing module, which may help other experts focus more on their unique domains of knowledge.

The abovementioned works make effort on optimizing the routing strategy of experts, while the router network is still the classical one that ignores the correlation between experts. Our work focus on designing the router network to incorporate the inherent correlation between experts. The routing network proposed in this paper is complementary to previous works.

## 3. Model Architecture

Yuan 2.0-M32 is based on the model structure of Yuan 2.0-2B (Wu et al., 2023). Yuan 2.0 introduces the local dependence of input tokens with the Localized Filtering-based Attention (LFA), to improve both the model's accuracy. In Yuan 2.0-M32, the dense feed-forward network (FFN) of every layer is replaced with a MoE component.

Figure 1 displays the architecture of MoE layer applied in our model. Taking four FFNs as an example (32 experts in fact), each MoE layer is composed of a group of individual FFNs as experts. The Router network ahead of experts dispatches the input token to the relevant expert(s). The classic Router network essentially establishes a feature vector for each expert, and computes the dot product between input token and the feature vector of each expert to obtain the specific likelihoods between token and experts. The experts with the strongest likelihood are selected for activation and participate in subsequent calculations.

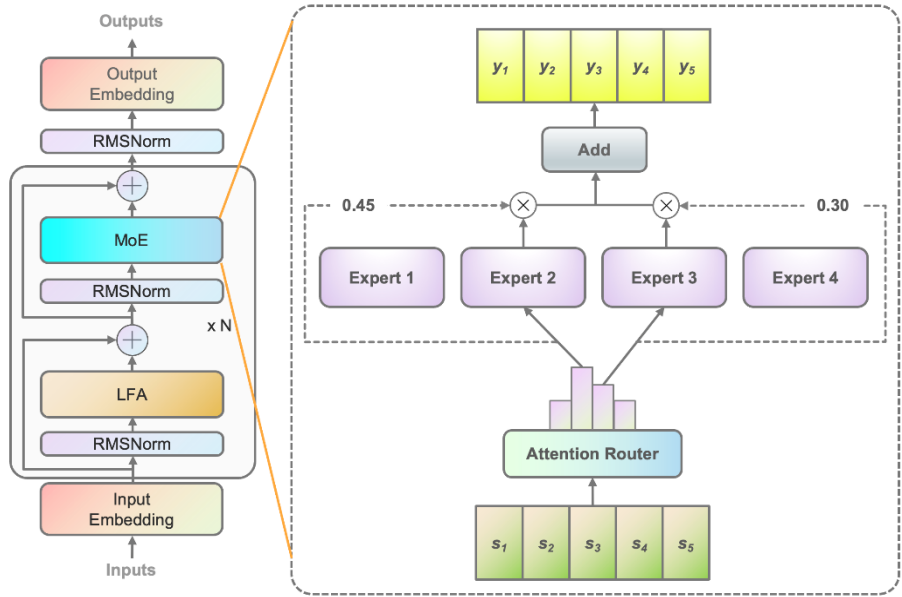

Figure 1: Illustration of Yuan 2.0-M32. Figure on the left showcases the scaling of Yuan 2.0 architecture with MoE layers. The MoE layer takes the place of the feed forward layer in Yuan 2.0. Figure on the right showcases the MoE layer structure. In our model, each input token will be assigned to 2 experts of the total 32, while in the figure we display 4 experts as an example. The output of the MoE is the weighted summation of the selected experts. N is the number of layers.

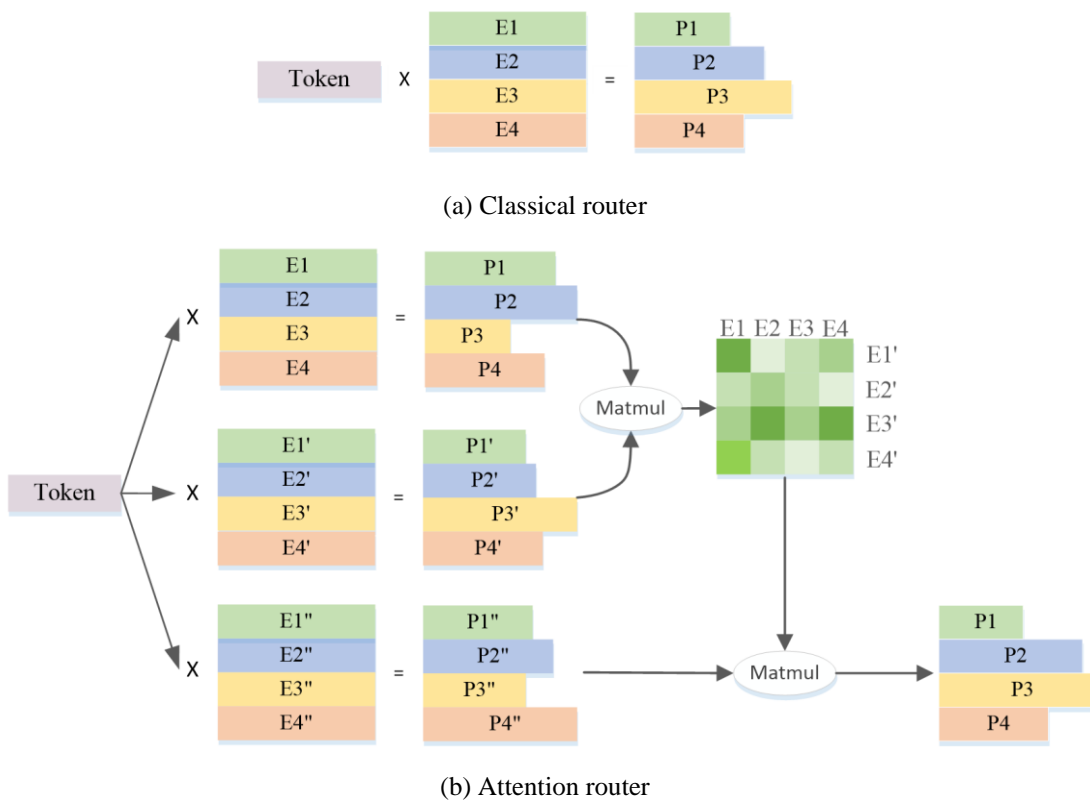

(a) Classical router

(b) Attention router

Figure 2: The overview of the attention router structure.

Figure 2(a) presents the structure of classical router network. The feature vectors of each expert are independent from each other, and the correlation between experts is ignored when calculating the probability. In fact, in most MoE models (Lepikhin et al., 2020; Fedus, Zoph and Shazeer, 2022; Zhou et al., 2022), two or more experts are usually selected to participate in the subsequent calculations, which naturally brings a strong correlation between experts. The consideration of correlation between experts will undoubtedly contribute to the improvement of accuracy.

Figure 2(b) presents the architecture of the Attention Router, a novel router network proposed in this work, incorporate correlation between experts by taking Attention mechanism. A coefficient matrix representing the correlation between experts is built, and then applied on the computation for the final probability value. In specific, given N experts for a token vector ($I \in R^d$), the expert routing process is as follows:

$$Q = WI, \quad W \in R^{N \times d}$$
$$K = W'I, \quad W' \in R^{N \times d}$$
$$V = W''I, \quad W'' \in R^{N \times d}$$
$$P = \text{Softmax}(QK^T)V, \quad P \in R^N$$

Then, the *M* experts are chosen by selecting top *M* values of P. In this paper, we set *M*=2, *N*=32, *d*=2048.

| Model | Params (M) | Test loss |
|---|---|---|
| Attention router | 826.0 | 2.109 |
| Classical router | 825.8 | 2.117 |
| Shared Expert router | 825.8 | 2.117 |

Table 1: Comparison of different router structures

Table 1 lists the accuracy results of different router. Our model is tested on 8 trainable experts with the Attention Router. The classical router model has 8 trainable experts to ensure a similar parameter scale, and the router structure is the same with that applied in Mixtral 8*7B (Jiang et al., 2024), which is a Softmax over a linear layer. The Shared Expert router takes the strategy of Shared Expert Isolation with classical router architecture (Dai et al., 2014). There are 2 fixed experts to capture the common knowledge and top-2 of 14 optional experts as the specialized ones. The output of MoE is the combination of the fixed and the ones selected by router. All the three models are trained with 30B tokens and tested with another 10B tokens. Considering the results between classical router and Shared Expert router, we find that the latter one gets exactly the same test loss with 7.35% more training time. The computational efficiency of the Shared Expert is relatively low, and it does not bring better training accuracy over the classical MOE strategy. Thus in our model, we take the classical routing strategy without any shared experts.

We test the scalability of the model by increasing number of experts and fixing the per-expert parameter size. The increase in the number of trainable experts only changes the model capacity, but not the actual activated model parameters. All the models are trained with 50B tokens and tested with another 10B tokens. We set the activated experts as 2, and the hyper-parameters for training are the same for the three models. The expert scaling effects is measured by the test loss after trained with 50B tokens (Table 2). Compared to the model with 8 trainable experts, model with 16 experts displays 2% lower loss, and model with 32 experts displays 3.6% lower loss. We choose 32 experts for Yuan 2.0-M32 considering its accuracy.

| Model | Test loss |
|---|---|
| 8 experts | 1.820 |
| 16 experts | 1.787 |
| 32 experts | 1.754 |

Table 2: Results of the scaling experiments

## 4. Training

**4.1 Model training**

Similar to the training strategy of Yuan 2.0, Yuan 2.0-M32 is trained with the combination of data parallelism and pipeline parallelism, however neither tensor parallelism nor optimizer parallelism is used. Training hyper-parameters are listed in Appendix A. Figure. 3 presents the loss curve, and the final training loss is 1.22.

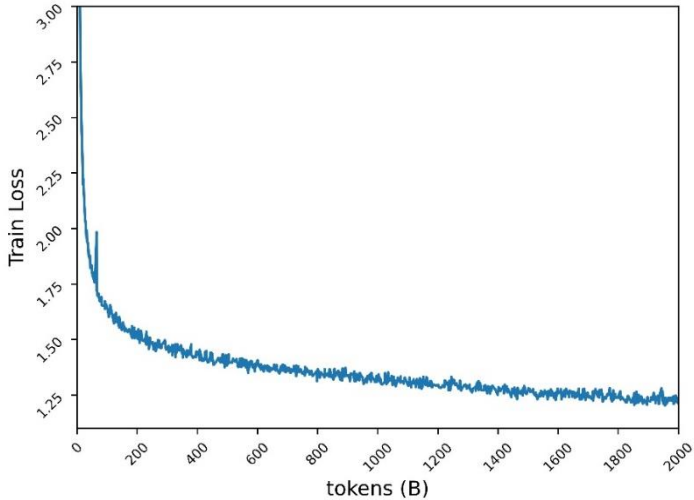

Figure 3: Pre-training loss of Yuan2.0-M32 on 2000B tokens

**4.2 Fine-tuning**

During fine-tuning, we extend the sequence length to 16384. Following the work of CodeLLama (Rozière et al., 2023), we reset the base value of Rotary Position Embedding (RoPE) frequencies to avoid the decay in attention scores with longer sequences. Instead of simply increasing the base value from 1000 to a much larger value (e.g. 1000000), we calculate the new base with the NTK-aware (bloc97, 2023), i.e.

$$b' = b \cdot s^{\frac{|D|}{|D|-2}}.$$

Where $b$ is the original base value ($b$=10000). $s$ is the number of extended times from the original context length to the extended context length. As we extend the context length from 4096 to 16384, s equals 4. $|D|$ is 128 in our setup. Therefore, the new base $b'$ is calculated to be 40890.

We also compare the performance of the pre-trained Yuan 2.0-M32 model with the NTK-aware styled new base, and with other base values (40000, 80000, 160000, 320000, 640000, 1280000, 2560000, 5120000, and 10240000) in the needle-retrieval task with sequence lengths up to 16K (gkamradt, 2023).

We find that the NTK-aware styled new base, 40890, performed better. Thus 40890 is applied during fine-tuning.

**4.3 Pre-training dataset**

Yuan 2.0-M32 is pre-trained with a bilingual dataset of 2000B tokens from scratch. The original data for pre-training contains more than 3400B tokens, and the weight for each category is adjusted according to the data quality and quantity.

The comprehensive pre-training corpus is composed of:

- 44 constituent sub datasets covering web crawled data, wiki, academic thesis, books, codes, math and formula, and domain-specific expertise. Some of them are open source datasets and the others created by Yuan 2.0.
- Parts of the common crawl data, Chinese books, dialogue and Chinese news data are inherited from Yuan 1.0 (Wu et al., 2021). Most of the pre-training data in Yuan 2.0 are also re-utilized.

Detailed information about the construction and source of each dataset is available below.

**Web**. (25.2%) Website crawling data is a collection from open source datasets and the common crawl data processed in our previous works (Yuan 1.0). Please refer to Yuan 1.0 for more details about the Massive Data Filtering System (MDFS) that extract contents in higher quality from web contexts

**Encyclopedia** (1.2%), **thesis** (0.84%), **book** (6.4%) and **translation** (1.1%) data are inherited from Yuan 1.0 and Yuan 2.0 datasets.

**Code** (47.5%). The code dataset is greatly expanded compared with Yuan 2.0. We adopt code from the Stack v2 (Lozhkov et al., 2024). The comments in Stack v2 are translated into Chinese. Code synthesized data created with the similar method as in Yuan 2.0.

**Math** (6.36%). All math data from Yuan 2.0 is reused. The data are predominantly from open source datasets, including proof-pile v1 (Azerbayev, 2022) and v2 (Paster et al., 2023), AMPS (Hendrycks et al. 2021), MathPile (Wang, Xia and Liu, 2023) and StackMathQA (Zhang, 2024). A synthetic dataset for numerical calculation is created with Python to benefit for four arithmetic operations.

**Specific-domain** (1.93%) is a dataset with knowledge from different background.

**4.4 Fine-tuning dataset**

The fine-tuning dataset is expanded based on the one applied in Yuan 2.0.

**Code Instruction dataset**. All the coding data with Chinese instruction and parts with English comments is generated with LLMs. About 30% of the code instruction data is in English, and the rest is in Chinese. The synthetic data are fabricated in a way that imitates the Python code with Chinese comments in terms of prompt generation and data cleaning strategy.

- Python code with English comments is collected from Magicoder-Evol-Instruct-110K (Wei et al., 2023) and CodeFeedback-Filtered-Instruction (Zheng et al., 2024). The instruction data with language tag such as "python" is extracted from the dataset, and organized into the format as shown in Appendix B. The dataset is also expanded with the Evol-instruct (Xu et al., 2023) and Self-instruct (Wang et al., 2022) method applied in the construction of Chinese Python code.

- Other codes such as C/C++/Go/Java/SQL/Shell etc., with English comments from open source dataset (Wei et al., 2023; b-mc2, 2023; Clinton, 2013; gayathrimanoj, 2023a, b; byroneverson, 2024; Zheng et al., 2024) are processed in a similar way with Python code. The cleaning strategies are similar to the method in Yuan 2.0. A sandbox is designed to extract compilable and executable lines in the generated codes, and keep the lines that pass at least one unit test.

**Math Instruction dataset**. The math instruction dataset are all inherited from the fine-tuning dataset in Yuan 2.0. To improve the ability of model to solve mathematical problems with programmatic methods, we construct a Program of Thoughts (PoT) prompting math data (Chen et al., 2022). PoT converts the mathematic problem into a code generation task that do calculations with Python.

**Safety Instruction dataset**. In addition to the chat dataset of Yuan 2.0, we construct a bilingual safe alignment dataset based on an open source safe alignment dataset (Ji et al., 2024). We only take the questions from the public dataset, and increase the variety of questions and regenerate Chinese and English answers with large language models.

**4.5 Tokenizer**

For Yuan 2.0-M32, the English and Chinese tokenizers are inherited from those applied in Yuan 2.0.

# 5. Results

We evaluate Yuan 2.0-M32 on Humaneval (Chen et al., 2021) for code generation, GSM8K (Cobbe et al., 2021) and the MATH (Hendrycks et al., 2021) for mathematical problem solving, ARC (Clark et al., 2018) for scientific knowledge and inference, and MMLU (Hendrycks et al., 2020) as an integrated benchmark.

**5.1 Code generation**

The capability of code generation is evaluated with the HumanEval Benchmark. The evaluation method and prompts are similar to those mentioned in Yuan 2.0, and the English prompted is constructed as Appendix B.

| Model | Params (B) | Active Params (B) | HumanEval (zero-shot) |
|---|---|---|---|
| Llama 3-70B | 70 | 70 | 81.7 |
| Llama 3-8B | 8 | 8 | 62.2 |
| Phi-3-medium | 14 | 14 | 62.2 |
| Phi-3-small | 7 | 7 | 61 |
| Phi-3-mini | 3.8 | 3.8 | 58.5 |
| Qwen1.5-72B | 72 | 72 | 68.9 |
| DeepseekV2 | 236 | 21 | 81.1 |
| Mixtral-8×22B | 141 | 39 | 45.1 |
| Mixtral-8×7B | 47 | 12.9 | 40.2 |
| Yuan 2.0-M32 | 40 | 3.7 | 74.4 |
| Yuan 2.0-M32 | 40 | 3.7 | 78.1 (14 shots) |

Table 3: Comparison of Yuan 2.0-M32 and other models on HumanEval pass@1.

The model is expected to complete the function after <sep>. And the generated function will be evaluated with unit tests. The results from zero-shot of Yuan 2.0-M32 and the comparison with other models are displayed in Table 3. The result of Yuan 2.0-M32 are second only to DeepseekV2 (DeepSeek-AI, 2024) and Llama3-70B (AI Meta, 2024), and far exceed the other models, even when its active parameters and computational consumptions are much lower than those from others. Compared with DeepseekV2, our model uses less than a quarter of the active parameters and less than a fifth of the computational effort per token, while reaching more than 90% of its accuracy level. And compared with llama3-70B, the gap between model parameters and computation is even greater, and we still reach 91% of its level. Yuan 2.0-M32 demonstrated reliable programming capability with three quarters of the questions passed. Yuan 2.0-M32 are good at few shot leaning. The accuracy of Humaneval is improved to 78.0 by taking 14 shots.

**5.2 Math**

The math capability of Yuan 2.0-M32 is evaluated with GSM8K and MATH benchmark. The prompts and test strategy for GSM8K is similar to that applied for Yuan 2.0, and the only difference is that we run it with 8 shots (Table 4).

| Model | Params (B) | Active Params (B) | GSM8K | MATH |
|---|---|---|---|---|
| Llama 3-70B | 70 | 70 | 93.0 | 50.4 |
| Llama 3-8B | 8 | 8 | 79.6 | 30 |
| Phi-3-medium | 14 | 14 | 91.0 | - |
| Phi-3-small | 7 | 7 | 89.6 | - |
| Phi-3-mini | 3.8 | 3.8 | 82.5 | - |
| Qwen1.5-72B | 72 | 72 | 81.9 | 40.6 |
| DeepseekV2 | 236 | 21 | 92.2 | 53.9 |
| Mixtral-8×22B | 141 | 39 | 78.6 | 41.8 |
| Mixtral-8×7B | 47 | 12.9 | 58.4 | 28.4 |
| Yuan 2.0-M32 | 40 | 3.7 | 92.7 | 55.9 |

Table 4: Comparison of Yuan 2.0-M32 and other models on GSM8K and MATH

MATH is a dataset with 12,500 challenging Mathematical Competition QA problems. Each question in this dataset has a complete step-by-step solution that leads the model to generate answer derivation and explanations. Answers to questions could be numerical values (0.5, 1/2, etc.), or mathematical expressions (y=2x+5, $x^2$+2x-1, 2a+b, etc.). Yuan 2.0-M32 produces the final answer with chain of thought (CoT) method with 4 shots. The answers will be extracted from analysis and transformed into a unified format. For numerical results, mathematically equivalent output in all formats are accepted. The answer of \frac{1}{2}, 1/2, 0.5, 0.50 are all converted into 0.5 and accepted as the same result. For mathematical expressions, we remove the tab and space symbol, and unified the regular expression of arithmetic operation. For instance, $y = (2x + 1)/5$, $y = \frac{2x+1}{5}$, $y = \frac{2x}{5} + \frac{1}{5}$, $y = 0.4x + 0.2$, …, etc., are all accepted as the same answers. The processed final results are compared with the ground truth answer, and evaluated with EM (exact match) scores.

From the results shown in Table 4, we can see that Yuan 2.0-M32 scores the highest on MATH benchmark. Compared to Mixtral-8×7B, which has 3.48 times larger active parameters than Yuan 2.0-M32, the score of Yuan is even nearly twice as high. On GSM8K, Yuan 2.0-M32 also achieves a score very close to that of Llama 3-70B, and outperforms other models.

5.3 **MMLU**

Massive Multitask Language Understanding (MMLU) covers 57 subjects in STEM, humanities, social sciences, etc., ranging from elementary language tasks to advanced logical inference tasks. All questions in MMLU are multi-choice QA questions in English. The model is expected to generate the correct option or corresponding analysis.

The input data for Yuan 2.0-M32 is organized as Appendix B. The text before <sep> is sent to the model, and all answer related to the correct answer or the option label is adopted as true.

The final accuracy is measured with MC1 (Table 5). The results on MMLU demonstrate the capabilities of our model in different domains. Yuan 2.0-M32 outperforms Mixtral-8×7B, Phi-3-mini, and Llama 3-8B in terms of performance.

| Model | Params (B) | Active Params (B) | MMLU |
|---|---|---|---|
| Llama 3-70B | 70 | 70 | 80.3 |
| Llama 3-8B | 8 | 8 | 68.4 |
| Phi-3-medium | 14 | 14 | 78.0 |
| Phi-3-small | 7 | 7 | 75.7 |
| Phi-3-mini | 3.8 | 3.8 | 68.8 |
| Qwen1.5-72B | 72 | 72 | 76.2 |
| DeepseekV2 | 236 | 21 | 77.8 |
| Mixtral-8×22B | 141 | 39 | 77.8 |
| Mixtral-8×7B | 47 | 12.9 | 70.6 |
| Yuan 2.0-M32 | 40 | 3.7 | 72.2 |

Table 5: Comparison of Yuan 2.0-M32 and other models on MMLU

5.4 **ARC**

AI2 Reasoning Challenge (ARC) benchmark is a multiple-choice QA dataset that contains questions from science exams through Grade 3 to 9. It is divided into Easy and Challenge parts, with the latter containing more complex parts that needs further reasoning. We test our model on the Challenge parts.

| Model | Params (B) | Active Params (B) | ARC-C |
|---|---|---|---|
| Llama 3-70B | 70 | 70 | 93.3 |
| Llama 3-8B | 8 | 8 | 78.6 |
| Phi-3-medium | 14 | 14 | 91.6 |
| Phi-3-small | 7 | 7 | 90.7 |
| Phi-3-mini | 3.8 | 3.8 | 84.9 |
| Qwen1.5-72B | 72 | 72 | 91.7 |
| DeepseekV2 | 236 | 21 | 92.3 |
| Mixtral-8×22B | 141 | 39 | 91.3 |
| Mixtral-8×7B | 47 | 12.9 | 85.9 |
| Yuan 2.0-M32 | 40 | 3.7 | 95.8 |

Table 6: Comparison of Yuan 2.0-M32 and other models on ARC-Challenge

The question and options are concatenated directly and separated with <n>, which is prompted as in Appendix B (similar to the pattern of MMLU). The text before <sep> is sent to model, and the model is expected to generate a label or corresponding answer. The generated answer is compared with the ground truth, and the results are calculated with MC1 target.

The results ARC-C are displayed in Table 6, and it shows that Yuan 2.0-M32 excels in solving complex scientific problems—it surpasses Llama3-70B in this benchmark.

| Model | Params (B) | Active Params (B) | GFlops per token (Inference) | GFlops per token (Fine-tune) | Average Accuracy | Average Accuracy/GFlops per token (Inference) |
|---|---|---|---|---|---|---|
| Llama 3-70B | 70 | 70 | 140 | 420 | 79.25 | 0.57 |
| Llama 3-8B | 8 | 8 | 16 | 48 | 64.15 | 4.00 |
| Qwen1.5-72B | 72 | 72 | 144 | 432 | 72.6 | 0.50 |
| DeepseekV2 | 236 | 21 | 42 | 126 | 79.05 | 1.88 |
| Mixtral-8×22B | 141 | 39 | 78 | 234 | 72.38 | 0.93 |
| Mixtral-8×7B | 47 | 12.9 | 25.8 | 77.4 | 60.83 | 2.36 |
| Yuan 2.0-M32 | 40 | 3.7 | 7.4 | 22.2 | 79.15 | 10.69 |

Table 7: Comparison of Yuan 2.0-M32 and other models on quality vs size. The mean accuracy is averaged on the scores of GSM-8K, Math, Humaneval, MMLU, and ARC-C.

From 5.1 to 5.4, we compare our performance to three MoE model (Mixtral family, Deepseek) and six dense models (Qwen (Bai et al., 2023), Llama family and Phi-3 family (Abdin et al., 2024)), to evaluate Yuan 2.0-M32's performance on different domains. Table 7 presents the comparison of Yuan 2.0-M32 with other models on accuracy vs computation. Yuan 2.0-M32 uses only 3.7B active parameters and 22.2 GFlops per token for fine-tuning, which is the most economical, to obtain comparable results or even surpass other models listed in the tables. Table 7 implies the outstanding computational efficiency and performance during inference of our model. The average accuracy of Yuan 2.0-M32 is 79.15 that is competitive with Llama3-70B. And the value of average accuracy / Glops per token is 10.69, which is 18.9 times larger than Llama3-70B.

## 6. Conclusion

In this work, we introduce Yuan 2.0-M32, a bilingual MoE language model based on Yuan 2.0. Attention Router that applied in this model achieves higher accuracy than classical router network. Yuan 2.0-M32 uses only 3.7B active parameters and 7.4 GFlops of inference per token, both of which are about 1/19 of Llama3-70B. In ARC-C benchmark, our model excels Llama 3-70B by 2.5 pts with only 5% active parameters. For the MATH benchmark, Yuan 2.0-M32 also achieves the highest score (55.9), surpassing Llama 3-70B by ~10% with ~5% computation cost. The results imply that our model has outstanding computational efficiency and performance during inference. We release our Yuan 2.0-M32 models at Github for public accessibility, as what we did for Yuan 2.0, and hope the open source model can benefit the development of LLMs and AI industry ecology.

**Appendix A: Hyper-parameters for Pre-training and fine-tuning**

| Parameter | Pre-train | Fine-tune |
|---|---|---|
| Learning rate (LR) | 1.0e-5 ~ 1.0e-4 | 8.0e-5 |
| LR decay style | cosine | constant |
| Sequence length | 4096 | 16384 |
| Global Batch size | 1536 | 1152 |

**Appendix B: Prompt examples for downstream tasks**

Code generation

---
Instruction: Given two positive integers a and b, return the even digits between a and b, in ascending order.

For example:
generate_integers(2, 8) => [2, 4, 6, 8]
generate_integers(8, 2) => [2, 4, 6, 8]
generate_integers(10, 14) => []
Response:
<sep>
```python
def generate_integers(a, b):
```
---

MMLU

---
Glucose is transported into the muscle cell:<n> A. via protein transporters called GLUT4. <n> B. only in the presence of insulin. <n> C. via hexokinase. <n>D. via monocarbylic acid transporters. <sep> A.

---

ARC-C

---
few-shot examples<n>question<n>optionA<n> optionB<n> optionC<n> optionD<sep> answer

---